# AI emotional support is better only when chosen, but shifts preferences even when it is not

Yaoxi Shi[1,2]

Cathy Mengying Fang[3]

Guy Laban[4]

Pattie Maes[3]

Amit Goldenberg[2,5,6]

[1]Imperial College Business School, Imperial College London
[2]Harvard Business School AI Institute, Harvard University
[3]MIT Media Lab, Massachusetts Institute of Technology
[4]Department of Industrial Engineering and Management, Ben-Gurion University of the Negev
[5]Harvard Business School, Harvard University
[6]Harvard Department of Psychology, Harvard University

**Corresponding Author:**
Yaoxi Shi: yaoxi.shi@imperial.ac.uk
Amit Goldenberg: agoldenberg@hbs.edu

**Abstract**

People increasingly face a novel decision when seeking emotional support: human or AI. In existing studies, AI's empathic messages are rated as well as or better than humans'. But these studies either assigned the support source or honored people's choice. In real life, support is often incongruent with choice, as people want one source and receive the other. Across three experiments (N = 1,951), participants chose whether to share an emotional experience with a human or an AI, then were randomly assigned to a congruent or incongruent partner. AI support was rated as superior only among those who had chosen it. Yet regardless of congruence, interacting with AI increased willingness to choose it again. In a 28-day study with OpenAI (N = 981), daily conversations shifted preferences toward AI and away from humans, but only when conversations turned personal. Emotional support choices are thus path-dependent, progressively redirecting away from human connection.

People are increasingly using generative AI for emotional support. On the user side, around 25% of people report using AI for emotional support[1]. On the platform side, AI companion applications such as Replika and Character.ai have attracted tens of millions of monthly active users. Industry reports from task-oriented AI platforms suggest that roughly 3% of conversations with general-purpose large language models qualify as emotional support[2,3]. Given the enormous volume of interactions on these platforms, the absolute number of such conversations is substantial. As the adoption of AI increases in every aspect of life, these numbers are likely to grow[4,5]. To understand this phenomenon, it is important to trace such behavior to its origins: the choice to turn to AI, both before initial use and after, which is the focus of the current paper.

In many cases, the choice of who to seek emotional support from is dependent on the availability of support sources. In most situations, people can choose to share their emotions with either humans or AI, and they receive emotional support from the source they prefer. In these cases, choice is driven by people's beliefs regarding the nature of the interaction[6]. To unpack how people make this choice, the first goal of the current paper is to understand people's beliefs about sharing emotions with a human and AI, and how these beliefs drive choice. In other situations, however, the desired source of support is unavailable, and people cannot choose who to seek emotional support from based on their preference. This is especially the case where people are seeking human support but turn to AI due to the limited availability and social frictions surrounding human support[1,7], although there are also cases in which people prefer AI but do not have access to AI support. The second goal of the paper is therefore to unpack the impact of choice congruence and incongruence on how people experience the AI or human emotional support they receive. Further, the increasing capability and adoption of AI in everyday life provide many opportunities for people to be exposed to AI emotional support, regardless of their choice[8,9]. Such direct experience with AI may update people's beliefs about AI's emotional support capability. Therefore, the third goal of the current paper is to understand how exposure to AI emotional support impacts subsequent choice on who to share emotions with.

**AI and Emotional Support**

Recent advances in generative AI have given AI systems the ability to engage in natural conversation with humans. Emerging evidence has suggested that, in the context of brief anonymous exchanges of emotional support messages, especially when providing empathic responses, AI often outperforms humans. Studies using third-party raters showed that raters evaluate empathic responses from AI as better than those from humans[10–14], even when the human responses are produced by doctors or professional support providers[10,13]. These results do not seem to be driven by a gap in supporters' motivation, and incentivizing people to write better responses does not reduce the gap between humans and AI[12]. Third-party rater studies have recently been extended to recipients' subjective ratings, showing again that even when it comes to participants' subjective ratings of the quality of interaction, AI is rated as better than human[12,15–18]. A recent meta-analysis that tried to summarize these findings estimated the effect of AI superiority in providing short empathic responses to be Cohen's d of *0.55*[19].

These findings, however, should be interpreted with caution, as most studies described above were conducted under conditions where participants were both unaware of the source of support and had no choice over it. Recent evidence suggests that source awareness can play a significant role[18,20]. Multiple studies on AI and human empathy suggest that labeling responses as AI-generated compared to human-generated reduces participants' satisfaction and perceived empathy[16,18,21]. This suggests that evaluations of empathy depend not only on the content of a response, but also on people's beliefs and preferences about who or what is providing it.

In everyday life, people typically choose where to seek emotional support before they receive it. This means that beliefs and preferences about the source of emotional support may be important factors shaping support seeking behavior. When it comes to choosing emotional support between a human and an AI, a few important aspects can shape people's decision. One obvious dimension is people's belief of the quality of emotional support provided by an AI compared to a human. While some believe that AI is superior to humans, others may think it is inferior[16,22–25]. A second important dimension seems to be the feeling of being judged[26–28]. Initial evidence from companion user interviews and surveys suggests that fear of judgment is a primary driver of choosing AI[1,29], and this belief towards humans and AI may also vary across individuals. To understand what shapes people's choices between human and AI, the first goal of the current paper is to map people's beliefs regarding human and AI emotional support and examine their association with choice.

Choice may also subsequently impact people's actual experience of the interaction. Research on choice outside the domain of empathic support suggests that receiving one's preferred option seems to drive more enjoyment and engagement, compared to receiving choice-incongruent options, even when choices are trivial or illusory[30–33]. In the context of AI emotional support, limited evidence points only to the comparison between human and AI when it is choice-congruent. In a recent study, participants were asked to choose whether to receive empathy from a human or an AI, and all received their choice-congruent options[17]. This was the first study where participants knew the source of support and still evaluated AI empathy as superior to humans. This finding suggests that receiving empathy from one's preferred source may meaningfully shape how its quality is experienced.

When it comes to choosing an empathic partner, choice congruence is not always guaranteed in practice. In everyday life, people often receive emotional support from sources not of their choice due to limited access[34]. An individual who would prefer emotional support from a human may have to turn to AI due to social isolation or merely lack of availability at the time they need such support[35]. Humans are inherently constrained by availability and social frictions. People may fail to find a human to seek emotional support from when they need it, or they may prefer a human but end up not seeking it due to fear of being judged or because it may affect their relationship[36]. In these situations, AI offers an accessible alternative, readily available at any moment with almost no cost[1,22]. Although less likely, the opposite is also possible: a person who wants to talk to an AI, to avoid judgment or because they don't perceive their social environment is capable of support, may not have access to AI. Therefore, understanding how receiving support not from one's desired source impacts the perceived quality of the interaction is important, especially as it becomes clearer that the use of AI for emotional support is primarily based on access, and those who use it often don't have access to human support[1]. The second goal of the current paper is therefore to understand how choice congruence and incongruence influence people's evaluation of the quality of interaction with either human or AI.

Assuming people receive emotional support from a certain source, whether it is human or AI, how does such exposure influence their future choices? A useful term to evaluate this aspect in decision-making literature is path dependence: the influence of previous experiences or choices on future choices. High path-dependence decisions are ones where previous experiences or choices highly influence future choices. One important question therefore is to what extent emotional support source choices are path-dependent, such that experiencing or choosing one source of support shifts the preference to that support in future choices. This is especially important for AI emotional support, because the widespread availability of AI means that exposure to AI emotional support often happens incidentally, without deliberate choice[37]. A large-scale analysis of discussions on Reddit's AI companionship community suggests that users do not encounter AI emotional support as a result of a deliberate search, but rather stumble upon it, primarily from having conversations with products like ChatGPT, Claude and Gemini[38]. Does exposure to AI emotional support shape people's beliefs about what AI can provide and shift their future preference? Given the ease of interaction and increasingly high-quality emotional support that AI is able to provide nowadays, such exposure may update people's perceptions of how empathic AI can be and increase their subsequent preference for choosing AI. The third goal of the current study is therefore to examine path dependence in emotional support choices.

**The Current Studies**

We used a combination of experimental and longitudinal studies to achieve three goals. The first goal was to understand people's beliefs of human and AI as emotional support sources and how they drive people's choice on who to seek emotional support from. The second goal was to examine how choice congruence and incongruence impact people's experienced quality of the emotional support. The third goal was to examine path dependence: how interacting with AI for emotional support impacts people's beliefs of AI and future choice.

To achieve these goals, we conducted a survey (Study 1), two experiments (Studies 2 and 3), and a longitudinal study (Study 4). Study 1 examined people's beliefs about sharing emotions with humans versus AI and how those beliefs shape their choices. Participants were asked to recall a recent significant emotional experience, rate their beliefs about sharing it with humans and AI on four dimensions: ability to provide emotional support, be non-judgmental, provide good advice and maintain confidentiality. Participants then chose to share their emotional experience with either an AI or a human. Studies 2 and 3 extended this design to test how choice congruence shaped the experienced quality of the interaction, and how the interaction itself reshaped beliefs about each partner and subsequent choices. After participants indicated their choice, they were randomly assigned to have a 5-minute conversation with either another human participant or an AI instructed to respond empathically (ChatGPT in Study 2 and Claude in Study 3), so the assignment was either congruent or incongruent with their initial choice. After the interaction, participants rated their actual experience with their assigned partner along the same four dimensions and rated the quality of the interaction. Finally, participants indicated who they would like to seek emotional support from in the future. Study 4, which was done in collaboration with OpenAI, which provided access to user conversation data, examined whether preference shifts following AI interaction extend to naturalistic, longitudinal settings. Over the course of a month, participants were instructed to engage in daily 5-minute interactions with an AI across three conversation topics (i.e., personal, non-personal, open-ended), and we investigated how these interactions with AI influenced their future preferences for talking with a human versus an AI about personal issues, and the role of different conversational behaviors in predicting the preference changes.

## Results

### Studies 1-3: Understanding Drivers and Consequences of Choosing Human or AI for Emotional Support

Studies 1-3 had a similar design for the pre-conversation stage, so we report them together below. Studies 1-3 recruited 1,951 participants in total (Study 1: 250; Study 2: 851; Study 3: 850). After preregistered exclusions, we have a sample of 1,894 for the pre-conversation analysis (Study 1: 229; Study 2: 842; Study 3: 823). Studies 2 and 3 extended the design of Study 1 to a real-time conversation between either human or AI, and they were identical in design. Study 3 was designed to replicate the results of Study 2 with a different large language model (Claude Opus 4.6 versus GPT-4o). In Studies 2 and 3, 596 participants were assigned to interact with an AI (Study 2: 287; Study 3: 309) and 1,069 were assigned to interact with a human (Study 2: 555; Study 3: 514). The human condition required twice as many participants because each interaction involved two people: one in the sharer role and one in the listener role. Because our primary analyses focused on participants in the sharer role, we excluded listeners from the human condition in our analysis. After preregistered exclusions, the final sample consisted of 944 (Study 2: 487; Study 3: 457) participants: 415 (Study 2: 221; Study 3: 194) in the human condition and 529 (Study 2: 266; Study 3: 263) in the AI condition (see full breakdown of exclusions and samples in Methods). We report the conversation and post-conversation results from Studies 2 and 3 together, which shared similar designs but used different large language models for the AI partner (ChatGPT in Study 2 and Claude in Study 3).

The results reported below follow the preregistered hypotheses for Study 3 (See preregistration at:
https://osf.io/9hsyc/overview?view_only=cca34e8db5da4db9b7f8092c9e2a6cf3), which were informed by results that emerged from Study 1 and Study 2 (See Supplementary Information section 1 for hypotheses comparison between the preregistrations). Study 3 was designed to test and validate those initial findings with a different large language model. Below, we present results that combine data from all relevant studies (see Supplementary Information section 2 for hypothesis testing within each study separately). For the regression analysis, we fitted mixed-effects models to the combined dataset including all relevant studies. Each model included a random intercept for study, allowing the baseline outcome level to vary across studies and thereby accounting for between-study heterogeneity.

## Experimental Design

**Pre-Conversation.** Participants were first asked to recall a recent significant emotional experience they had in the past month, either positive or negative (see Methods for specific instructions), and rate the emotions in the experience (see Supplementary Information section 3 for further analysis). Then, they rated their beliefs of both sharing the experience with an AI chatbot and a human listener along four dimensions: emotional support, nonjudgment, confidentiality, and advice quality (see Supplementary Information section 4 for factor analysis). Participants also rated the importance of each factor in choosing a sharing partner (see Supplementary Information section 5 for further analysis). Then participants made the choice whether they preferred to talk to an AI chatbot or a human participant using both a binary choice question and continuous rating scales for each option. The order of belief rating and choice was reversed between Studies 1 and 2-3 to rule out the possibility that choice may influence people's beliefs (see Methods for a description of the order).

**Conversation.** Study 1 ended after participants provided their choice and rated their beliefs. Study 2 (ChatGPT) and 3 (Claude) were identical in design and extended Study 1 with a real interaction with either a human or an AI. After making their choice, participants in Studies 2 and 3 were then randomly assigned to engage in a 5-minute conversation via text messages about their recalled experiences with either an AI chatbot (Study 2: powered by GPT-4o; Study 3: powered by Claude Opus 4.6), or another human participant acting as a listener. Assignments to human and AI were either congruent or incongruent with participants' choice. Participants were informed of their assigned partner before the conversation began. Both the AI chatbot and human listener received the same instructions, which were "to offer a supportive, empathetic response throughout the conversation" (see Methods for the complete instruction). The structure of the conversation was consistent across conditions. The AI chatbot or the human listener started the conversation with a greeting, then participants were instructed to share the emotional experience that they recalled at the beginning of the survey, after which the AI chatbot or the human listener responded and continued chatting by exchanging messages for a total of five minutes. Participants were not allowed to finish the conversation before the time ended but could slow down their response rate or stop responding if they wanted (as preregistered, we excluded participants for whom either they or their partner sent fewer than three messages during the conversation, see Methods for this exclusion).

**Post-Conversation.** After the conversation, participants evaluated their experience of sharing the emotional experience with their assigned partner on the same four dimensions (i.e., emotional support, nonjudgment, confidentiality, and advice quality). This was to examine how the interaction experience changed participants' beliefs of sharing emotions with a human or an AI chatbot. They then reported their experienced empathy, enjoyment of and satisfaction with the conversation, as well as their desire to continue the conversation. They also indicated their future preference for sharing significant emotional experiences with a human or an AI, using both a binary choice question and continuous rating scales for each option. Finally, participants answered their general experience with and attitudes of AI and their demographics.

## Pre-Conversation Beliefs and Choice

The first goal of this paper was to understand people's beliefs of sharing emotions with a human and an AI, and how these beliefs influence their choice on who to share with. We hypothesized that people generally perceive humans as more judgmental but also more emotionally supportive than AI (H1a). However, we also expected these general perceptions to diverge depending on their choice. Specifically, we hypothesized that people who choose to share with an AI perceive AI as less judgmental than a human but not less emotionally supportive (H1b). Conversely, people who choose to share with a human perceive humans to be more emotionally supportive than AI but not more judgmental (H1c).

**How do people perceive sharing emotional experiences with a human and an AI?** In Studies 1-3, participants rated their beliefs of sharing emotions with a human partner and an AI chatbot across 14 measurement items. These 14 items were clustered into four dimensions using factor analysis: emotional support, perceived judgment, confidentiality, and advice quality (See Methods for specific items and Supplementary Information section 4 for factor analysis). To test H1a, we conducted two-sided paired-samples t-tests comparing participants' ratings of human and AI partners on perceived emotional support and perceived judgment. Consistent with H1a, results indicated that human partners were perceived as offering greater emotional support than AI (Studies 1-3 combined: $t(1893) = 18.62$; $P < 0.001$; Cohen's $d = 0.43$; mean difference, 0.79; 95% CI, (0.71, 0.88); Fig. 1a), but also as more judgmental ($t(1893) = 31.11$; $P < 0.001$; Cohen's $d = 0.71$; mean difference, 1.32; 95% CI, (1.24, 1.40)). We additionally examined the other two dimensions: perceptions of confidentiality and advice quality. Participants perceived sharing emotions with AI as more confidential than sharing with a human partner ($t(1893) = 7.53$; $P < 0.001$; Cohen's $d = 0.17$; mean difference, 0.39; 95% CI, (0.29, 0.49)). Perceived ability to provide advice did not differ significantly between AI and human partners ($t(1893) = 0.33$; $P = 0.743$; Cohen's $d = 0.01$; mean difference, 0.01; 95% CI, (−0.07, 0.09)).

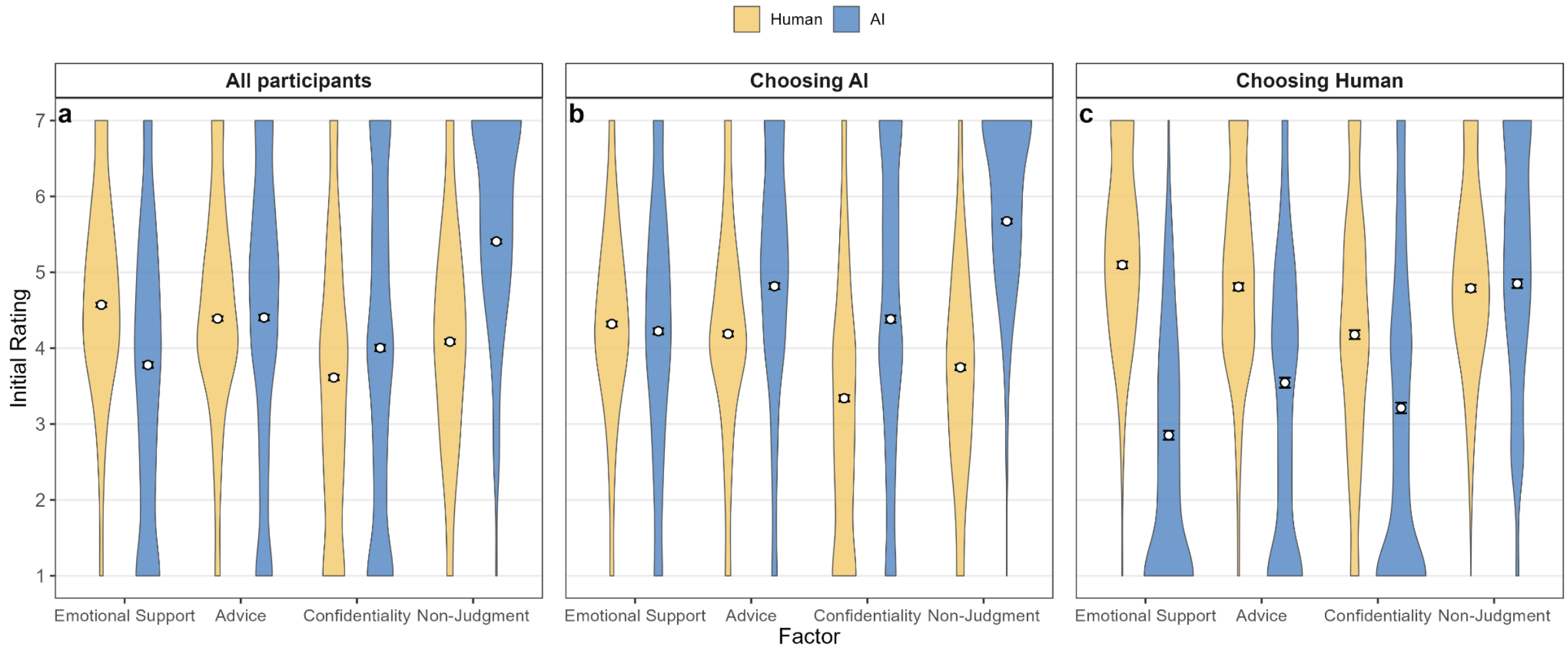

**Fig. 1 | Beliefs of sharing emotional experience with an AI vs. a Human before conversation, overall and by initial sharing choice (Studies 1-3 combined).** (**a**) Across all participants, humans were perceived as providing more emotional support and as more judgmental than AI. AI was perceived as more confidential. Advice quality did not differ significantly. (**b**) Among participants who initially chose AI, AI was perceived as less judgmental, more confidential, and offering higher-quality advice, and humans were perceived as providing slightly more emotional support than AI. (**c**) Among participants who initially chose a human partner, AI and human partners were rated as similar in nonjudgment, but humans were rated as higher in emotional support and advice quality. Error bars represent standard errors of the mean.

**How do beliefs predict choice?** To examine the relative importance of participants' beliefs in shaping people's choice, we entered the difference scores between ratings of humans and AI on four perception dimensions (i.e., perceived nonjudgment, emotional support, confidentiality, and advice quality) into a logistic regression model predicting participants' binary choice of conversation partner. Table 1 reports the regression results. Comparing magnitudes of the four predictor coefficients, the emotional support difference score showed the largest effect, closely followed by the nonjudgment difference score, suggesting that both nonjudgment and emotional support were the primary predictors of participants' choice of whom to talk to.

**Table 1 | Logistic Regression Predicting Preference for the Human Option from Human–AI Perception Difference Scores (Studies 1-3 Combined)**

| Predictor | β (SE) | Odds ratio (95% CI) | P |
|---|---|---|---|
| Emotional support | 1.18 (0.11) | 3.25 (2.62, 4.03) | < 0.001 |
| Nonjudgment | 0.81 (0.10) | 2.25 (1.86, 2.73) | < 0.001 |
| Advice quality | 0.34 (0.11) | 1.40 (1.13, 1.73) | 0.002 |
| Confidentiality | 0.30 (0.09) | 1.36 (1.14, 1.61) | < 0.001 |

**Who did participants choose?** In all Studies 1-3, participants were asked to choose whether they would prefer to share the recalled emotional experience with an AI chatbot or with a human partner. Results suggested that a majority of participants (Study 1: 75.5%; Study 2: 69.6%; Study 3: 63.2%) chose to share their emotional experience with the AI chatbot rather than with a human. These choices were different from previous studies[17], which showed a higher preference in choosing a human partner for receiving empathy. One potential reason for this difference is that Wenger et al. (2026) involved only one exchange, whereas this study involved a conversation, which may have made expected judgment from a human more salient and shifted the preferences towards AI. In our own Study 4 reported below, we also observed that more participants preferred talking to a human about personal matters in the pre-study survey, perhaps because the question did not specify that the human option referred to a stranger, as it did in Studies 1-3. Participants may have construed

"human" more broadly to include close others or a therapist, which could increase the likelihood of choosing the human option. Either way, we do not think of these percentages as the main topic of investigation, but rather focus on the drivers and consequences of interacting with a congruent or incongruent partner.

In addition to this binary choice, participants also reported their comfort (Study 1; from 0: Not at all to 10: Very much) and willingness (Studies 2 and 3; from 0: Not at all to 7: Very much) to share their emotional experiences with each option using continuous rating measures. A two-sided paired samples t-test indicated that participants reported significantly greater comfort and willingness to share with the AI chatbot than with a human partner (standardized: $t(1893) = 15.80$; $P < 0.001$; Cohen's $d = 0.36$; mean difference, 0.51 SD; 95% CI, (0.45, 0.57)). These preferences on who to share with were strongly associated with their emotion sharing tendencies with humans and AI in everyday life (See Supplementary Information section 7a for results).

**How did participants' beliefs differ conditional on choice?** Interestingly, we found that participants' beliefs regarding sharing emotions with human and AI differed systematically as a function of their choice on who to share with. Among participants who chose to share their emotional experience with the AI, two-sided paired-samples t tests indicated that they perceived a human as providing slightly higher emotional support than AI ($t(1278) = 2.22$; $P = 0.026$; Cohen's $d = 0.06$; mean difference, 0.10; 95% CI, (0.01, 0.18); Fig. 1b). This is inconsistent with H1b, in which we hypothesized no difference; the difference is very small. They perceived AI as significantly less judgmental than a human partner ($t(1278) = -41.80$; $P < 0.001$; Cohen's $d = -1.17$; mean difference, -1.93; 95% CI, (-2.02, -1.84)). These participants also perceived sharing with AI as more confidential ($t(1278) = 18.25$; $P < 0.001$; Cohen's $d = 0.51$; mean difference, 1.04; 95% CI, (0.93, 1.16)), and perceived AI as better able to provide advice ($t(1278) = 15.23$; $P < 0.001$; Cohen's $d = 0.43$; mean difference, 0.63; 95% CI, (0.55, 0.71)).

Participants who chose to share their emotional experience with the human partner revealed a completely different set of beliefs regarding human and AI empathic support. Two-sided paired samples t-tests indicated that they perceived sharing with a human as equally non-judgmental as sharing with AI ($t(614) = 0.96$; $P = 0.335$; Cohen's $d = 0.04$; mean difference, 0.06; 95% CI, (-0.06, 0.19); Fig. 1c). Human partners were perceived as significantly more emotionally supportive ($t(614) = 34.70$; $P < 0.001$; Cohen's $d = 1.40$; mean difference, 2.24; 95% CI, (2.12, 2.37)) and better able to provide advice ($t(614) = 18.49$; $P < 0.001$; Cohen's $d = 0.75$; mean difference, 1.26; 95% CI, (1.13, 1.40)), and more confidential ($t(614) = 11.54$; $P < 0.001$; Cohen's $d = 0.47$; mean difference, 0.97; 95% CI, (0.80, 1.13)) than the AI partner, consistent with H1c. These results were robust in each individual study (see Supplementary Information section 2 for results).

Across the three studies, we show that people who chose to share with an AI held systematically different beliefs of humans and AI than those who chose to share with a human. Those who chose AI perceived it as providing emotional support comparable to that of humans and perceived AI as better on perceived judgment, advice quality, and confidentiality. Conversely, those who chose a human perceived humans as equally nonjudgmental as AI, and perceived a human as better than AI on perceived emotional support capabilities, advice quality, and confidentiality.

**Experienced Quality of the Conversation**

The second goal of the current paper was to understand how the congruence between participants' choice and their actual assigned conversation partner influenced the experienced quality of the conversation, and whether this differed for people who interacted with a human versus an AI. In both Studies 2 and 3, participants were randomly assigned to engage in a 5-minute conversation with either an AI chatbot (powered by GPT-4o in Study 2 and powered by Claude Opus 4.6 in Study 3) or a human listener, independent of their initial preference. Participants engaged in a 5-minute conversation with their assigned partner (see Methods for descriptives of the conversations). We hypothesized that choice congruence would have a positive main effect on people's sharing experience (H2a).

**How does the congruence between choice and chat condition influence conversation outcomes?** We conducted linear regression models predicting participants' experienced empathy, desire to continue the conversation, experienced satisfaction and experienced enjoyment of the conversation from a binary indicator of choice–assignment congruence (0 = incongruent, 1 = congruent). We found that choice–assignment congruence had a positive main effect on all these outcomes. Participants whose assigned partner matched their initial choice reported higher experienced empathy ($t(942) = 3.79$; s.e. = 0.03; $P < 0.001$; $\beta = 0.12$; 95% CI, (0.06, 0.19)), a stronger desire to continue the interaction ($t(942) = 6.25$; s.e. = 0.03; $P < 0.001$; $\beta = 0.20$; 95% CI, (0.14, 0.26)), higher enjoyment ($t(942) = 5.40$; s.e. = 0.03; $P < 0.001$; $\beta = 0.17$; 95% CI, (0.11, 0.24)) and satisfaction with the conversation ($t(942) = 5.66$; s.e. = 0.03; $P < 0.001$; $\beta = 0.18$; 95% CI, (0.12, 0.24)). These main effects were, however, moderated by the actual conversation partner. Congruence had positive effects on the four outcomes among participants assigned to chat with an AI ($Ps < 0.001$), but had no significant effect among those assigned to a human partner ($Ps > 0.40$; see Supplementary Information section 2 for more detailed results).

This asymmetry reveals an interesting finding: AI emotional support was rated more favorably than human support only when participants had initially chosen AI. Participants who initially chose AI and chatted with AI evaluated the interaction better than those who chose AI but chatted with a human (empathy: $t(614) = -6.12$; s.e. = 0.12; $P < 0.001$; b = -

0.72; 95% CI, (-0.95, -0.49); enjoyment: t(614) = -7.99; s.e. = 0.14; P < 0.001; b = -1.09; 95% CI, (-1.36, -0.83); willingness to continue: t(614) = -7.40; s.e. = 0.16; P < 0.001; b = -1.15; 95% CI, (-1.46, -0.85); satisfaction: t(614) = -9.27; s.e. = 0.13; P < 0.001; b = -1.22; 95% CI, (-1.48, -0.96)) and those who chose a human and chatted with a human (empathy: t(513) = -5.61; s.e. = 0.14; P < 0.001; b = -0.78; 95% CI, (-1.06, -0.51); enjoyment: t(513) = -6.74; s.e. = 0.16; P < 0.001; b = -1.09; 95% CI, (-1.41, -0.77); willingness to continue: t(512.8) = -5.26; s.e. = 0.19; P < 0.001; b = -1.00; 95% CI, (-1.37, -0.62); satisfaction: t(513) = -8.52; s.e. = 0.15; P < 0.001; b = -1.30; 95% CI, (-1.60, -1.00)). The ratings of participants who chatted with AI but chose a human were not significantly higher than any of the human assignment conditions (Ps > 0.05), except that they were more satisfied than those who chose AI and chatted with a human (t(427) = -2.45; s.e. = 0.18; P = 0.015; b = -0.44; 95% CI, (-0.78, -0.09)) and those who chose a human and chatted with a human (t(326) = -2.55; s.e. = 0.20; P = 0.011; b = -0.52; 95% CI, (-0.92, -0.12); this comparison became significant only when we aggregated the data from both studies). See Fig. 2. These results point to two important aspects. First, the effect of congruence is primarily driven by those who chose and received AI. Second, and perhaps more important, is the fact that AI superiority over humans in short empathic communication is primarily driven by participants who initially preferred AI. This suggests an important moderator for the AI superiority findings.

**Conversation analysis.** An important question raised by these findings is whether choice congruence affected ratings of AI empathy by changing the actual content of the conversation or merely participants' interpretation of it. To examine this question, we used an LLM-as-a-judge approach to code the level of empathy in the AI's responses (see Supplementary Information section 14 for the specific method). We found no significant differences on the level of empathy AI provided in the conversations for participants who received a choice-congruent versus an incongruent partner (t(525) = 0.02; s.e. = 0.02; P =

0.98; β = 0.0005; 95% CI, (−0.04, 0.04)). This suggests that the effect of choice congruence influenced participants' subjective experiences of AI empathy, rather than the objective differences in the AI's responses.

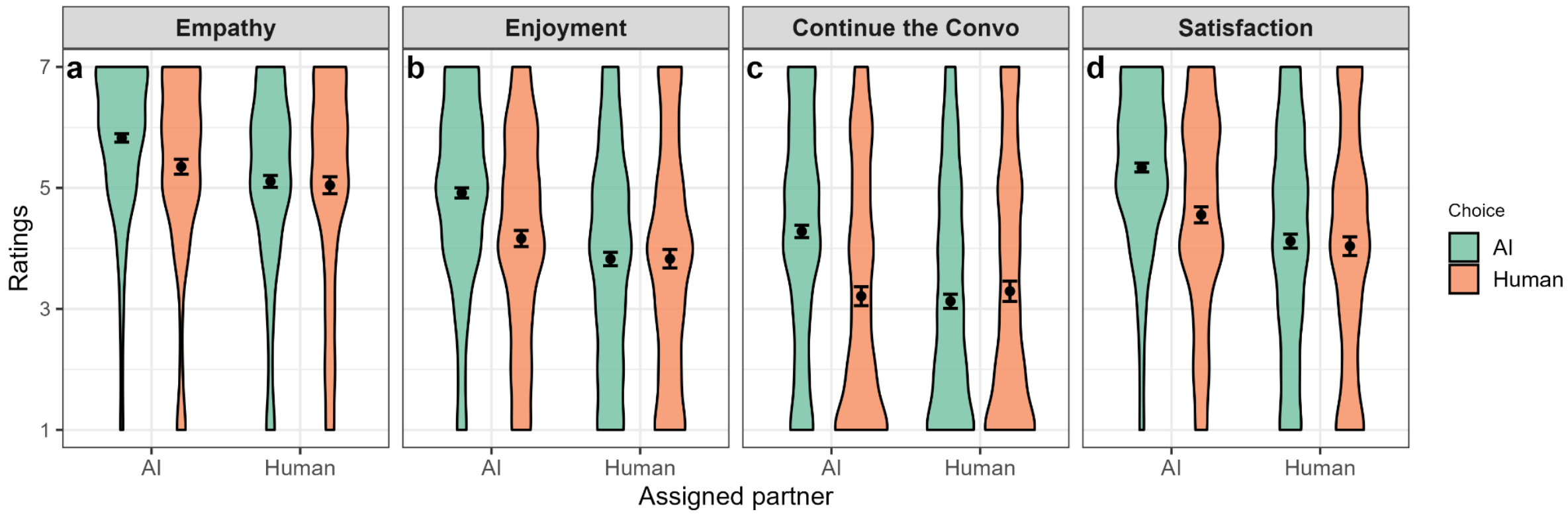

**Fig. 2 | Effects of choice and assigned partner on reported conversation experiences, combined results from Studies 2 and 3.** Choice × partner interactions were significant for experienced empathy (**a**), enjoyment (**b**), desire to continue (**c**), and satisfaction (**d**). Error bars represent standard errors.

## Post-Conversation Perception and Choice

The third goal of the paper was to examine how the actual experience of sharing emotions with an AI or a human changed participants' beliefs about their assigned partner and influenced their choice on who to share emotions with in the future. After the conversation, participants rated their post-conversation beliefs of their assigned partner on the same four dimensions as before (i.e., emotional support, nonjudgment, advice quality, and confidentiality). They then indicated whether they would prefer an AI or a human stranger for future emotional sharing. We hypothesized that people assigned to share emotional experience with an AI (vs. a human) would be more likely to choose to share emotional experiences with an AI in the future, regardless of their initial choice (H3).

**How does the interaction change people's beliefs on sharing emotions with a human and an AI?** We compared participants' pre- and post-conversation ratings of their assigned partner across the four dimensions. In the results below, we focus on perceived emotional support and nonjudgment, as these two dimensions showed the largest human-versus-AI differences and were the strongest predictors of participants' initial sharing choices. (See Supplementary Information section 9 for results on advice quality and confidentiality).

We found that the 5-minute interaction with an AI or a human significantly changed people's beliefs about sharing emotions with their assigned partner. After sharing with an AI, two-sided paired sample t-tests showed that, regardless of the initial choice, participants perceived the AI as providing more emotional support ($t(528) = 21.00$; $P < 0.001$; Cohen's $d = 0.91$; mean difference, 1.07; 95% CI, (0.97, 1.17); Fig. 3a) and being more nonjudgmental ($t(528) = 11.51$; $P < 0.001$; Cohen's $d = 0.50$; mean difference, 0.52; 95% CI, (0.43, 0.61); Fig. 3b) than their pre-conversation beliefs. Interestingly, among participants who initially chose to share with a human, chatting with an AI raised their beliefs of AI's emotional support to the same level as the pre-conversation beliefs of participants who had initially chosen AI ($t(299.7) = -0.41$; $P = 0.685$; Cohen's $d = -0.04$; mean difference, -0.06; 95% CI, (-0.35, 0.23)).

However, results for human partners were mixed. After sharing with a human, participants perceived human partners as more nonjudgmental than their pre-conversation beliefs ($t(414) = 20.80$; $P < 0.001$; Cohen's $d = 1.02$; mean difference, 1.23; 95% CI, (1.11, 1.34); Fig. 3d). But the perceived human emotional support did not change in a positive way ($t(414) = -1.76$; $P = 0.078$; Cohen's $d = -0.09$; mean difference, -0.15; 95% CI, (-0.31, 0.02)). Those who initially chose to share with a human actually reported receiving less emotional support from their human partner than they had anticipated ($t(156) = -4.82$; $P < 0.001$; Cohen's $d = -0.38$; mean difference, -0.69; 95% CI, (-0.98, -0.41); Fig. 3c). Those who initially chose to share with an AI reported emotional support from the human partner that was similar to their pre-conversation beliefs (t(257) = 1.87; P = 0.062; Cohen's d = 0.12; mean difference, 0.18; 95% CI, (-0.01, 0.38)).

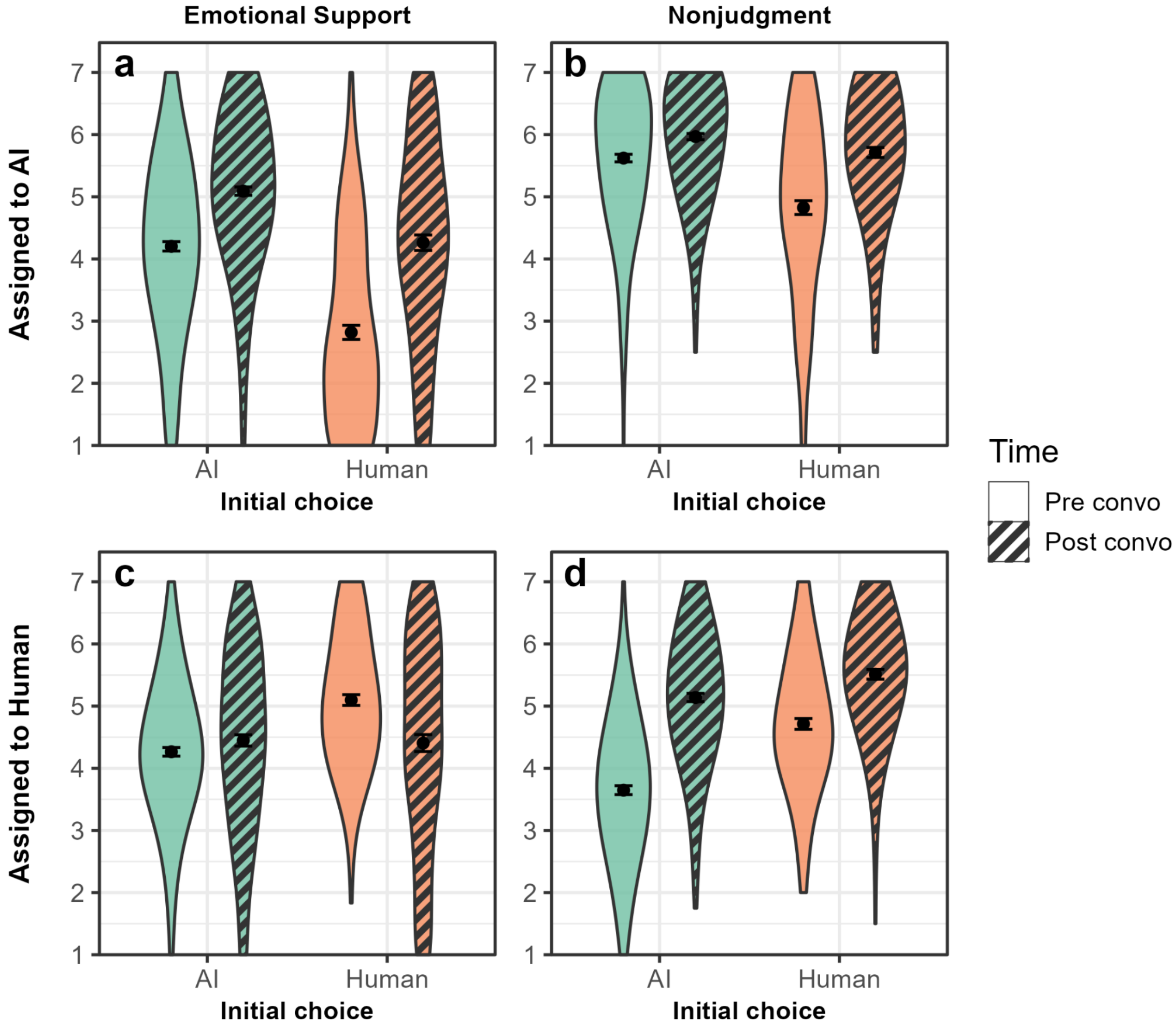

**Fig. 3 | Pre- to post-conversation changes in perceived emotional support and nonjudgment of the assigned partner by initial sharing choice, combined results from Studies 2 and 3. a, b,** For participants assigned to chat with an AI, they reported experiencing more emotional support (**a**) and AI as more nonjudgmental (**b**) than their pre-conversation beliefs. **c, d,** For participants assigned to chat with a human, overall they reported experiencing similar levels of emotional support as their pre-conversation beliefs (**c**) but experienced humans as more nonjudgmental (**d**). Error bars represent standard errors.

**How does the interaction change people's future choice on who to share emotions with?** After the conversation, participants were asked to choose whether they would prefer to share a future significant emotional experience with an AI or a human stranger. We hypothesized that people who shared emotional experience with an AI (vs. a human) would be more likely to choose AI in the future (H3). As preregistered, we conducted a logistic regression model predicting participants' future choice using their assigned conversation partner in the study. Participants who had chatted with an AI were significantly more likely to choose an AI partner for their future emotion sharing than those who had chatted with a human ($z = 7.54$; s.e. = 0.07; $P < 0.001$; $\beta = 0.51$; 95% CI, (0.38, 0.65)). In fact, 70.1% of participants assigned to chat with an AI chose to share emotions with an AI (vs. a human) in the future, compared to 45.5% in the human condition. This result suggests that a 5-minute interaction of talking about emotions with an AI (vs. a human) creates a substantial difference in people's preference on who to share emotions with in the future. (See Supplementary Information section 11 for results on the continuous measures of future sharing with an AI and a human.)

This main effect of chat condition is consistent across participants who initially chose to share with either a human or an AI. For participants who initially chose to share with AI, being assigned to chat with a human (vs. an AI) predicted a higher likelihood of choosing a human in the future ($z = 6.64$; s.e. = 0.11; $P < 0.001$; $\beta = 0.70$; 95% CI, (0.50, 0.91); see Fig. 4a). For participants who initially chose to share with a human, being assigned to chat with an AI (vs. a human) also significantly predicted a higher likelihood of choosing an AI in the future ($z = 4.05$; s.e. = 0.15; $P < 0.001$; $\beta = 0.61$; 95% CI, (0.31, 0.90); see Fig. 4b). These results suggest that the choice of emotional support partner is path-dependent: an actual experience of sharing emotions with a certain partner increases the likelihood of choosing to share with the same partner in the future.

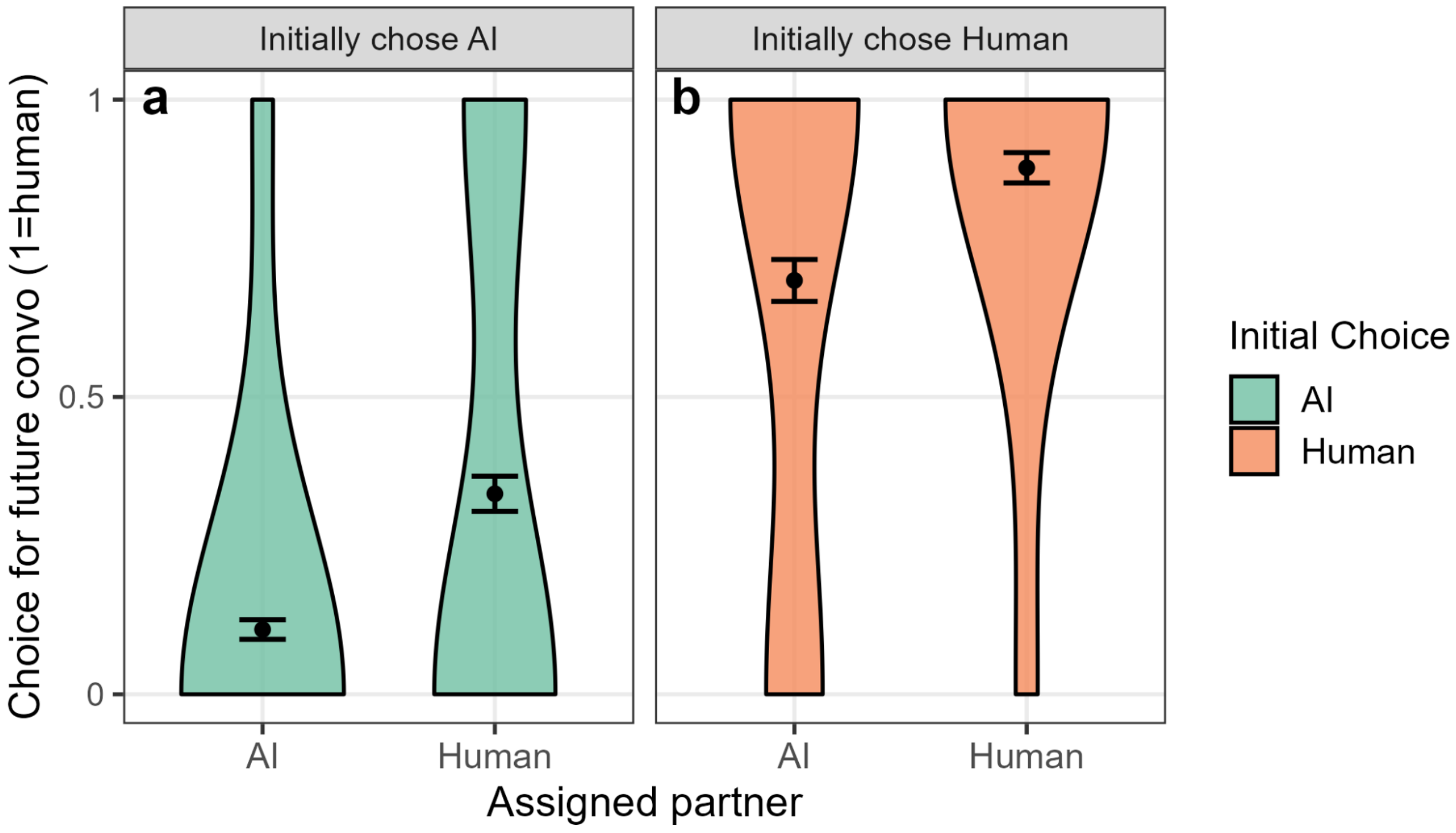

**Fig. 4 | Future sharing preference after the conversation, by initial choice and assigned partner, combined results from Studies 2 and 3. a,** Among participants who initially preferred AI, being assigned to chat with a human (vs. AI) increased the likelihood of choosing a human for future sharing. **b**, Among participants who initially preferred to share with a human, being assigned to chat with AI (vs. a human) increased the likelihood of choosing AI for future emotional sharing. The y axis depicts the proportion choosing a human for future sharing. Error bars represent standard errors.

**Study 4: Examining Effects of Using AI on Future Preference in a Longitudinal Study**

Experiments 2 and 3 showed that the experience of sharing emotions with an AI significantly increased the probability of choosing to share with an AI instead of a human in the future. However, the design had a few limitations. First, participants were asked to make their choice immediately after the interaction, so the preference shift they reported may reflect a recency effect rather than a durable change in preference. Second, it is unclear whether having the same choice of partner, particularly in the incongruent conditions, is driven by a strategic consideration where people hope to receive the opposite of what they asked. More broadly, the experiments tested the effect of a single structured short interaction in a laboratory setting, yet in everyday life, people engage with AI chatbots in varied ways and over extended periods. It remains unclear whether the observed effect would hold in such real-world contexts.

To address these limitations, we conducted a longitudinal naturalistic experiment in collaboration with OpenAI in Study 4 to examine how interacting with an AI daily over the course of a month changes people's preferences. Rather than focusing only on emotional experiences as in Studies 1-3, and to accommodate other research goals, in this study we asked participants to engage with AI across a broader range of personal topics, which better reflects how people use AI in everyday life. We also aimed to examine how the content of the interaction matters for preference change by randomly assigning participants to engage with AI about personal issues, non-personal issues, and open-ended conversations.

**Experimental Design**

We conducted a 4-week (28-day) randomized controlled trial where participants were instructed to engage in at least a 5-minute chatbot conversation every day for 28 days with daily reminders. The study had a 3 × 3 design, meant to accommodate a few research goals that are unrelated to the current study. The first variable was *conversation topic*, where participants were asked to interact with ChatGPT every day about one of the three topics: personal topics, non-personal topics, and open-ended. Participants who were assigned to the non-personal or personal topic conditions received specific prompts they were to discuss with ChatGPT each day (See Supplementary Information section 13 for the full list of prompts). Participants assigned to the open-ended condition did not receive specific topic prompts; they were encouraged to talk about any topic they would like. The second variable was *modality*, in which participants were assigned to chat with ChatGPT only through text messages, or only through voice where the AI model was configured to respond with either a neutral voice or an expressive voice. *Modality* is irrelevant to the goal of the current paper; we report the effect of Modality on preference in Supplementary Information section 18, and other outcomes related to this variable are covered in a second paper (see Fang et al., 2025 for full reporting of the results)[39].

At the beginning of the study, participants filled out a survey reporting their baseline choice of conversation partner for talking about personal issues, among other scenarios. For each scenario, participants reported whether to choose AI or not, and whether to choose a human or not. Full details can be found in Methods. After 28 days, participants completed a post-study survey in which they reported their choices again on who to interact with for talking about personal issues, as well as the other scenarios in the pre-study survey measure. See Supplementary Information section 17 for the full results.

To examine whether the conversation topic manipulation produced the intended conversational content, we classified the conversation corpus using the LLM-as-a-judge[40] method. We first prompted GPT-4o to produce a one-sentence summary of the conversation contents, and then we used GPT-4o-mini to map the summary to a conversation topic category. Topic manipulation matched expectations. More than 40% of conversations in the *Personal* condition were classified as "Emotional Support & Empathy," compared with substantially lower rates in the *Open-ended* and *Non-personal* conditions. See Fig. 5. Full methods and analyses can be found in Supplementary Information sections 16 and 19.

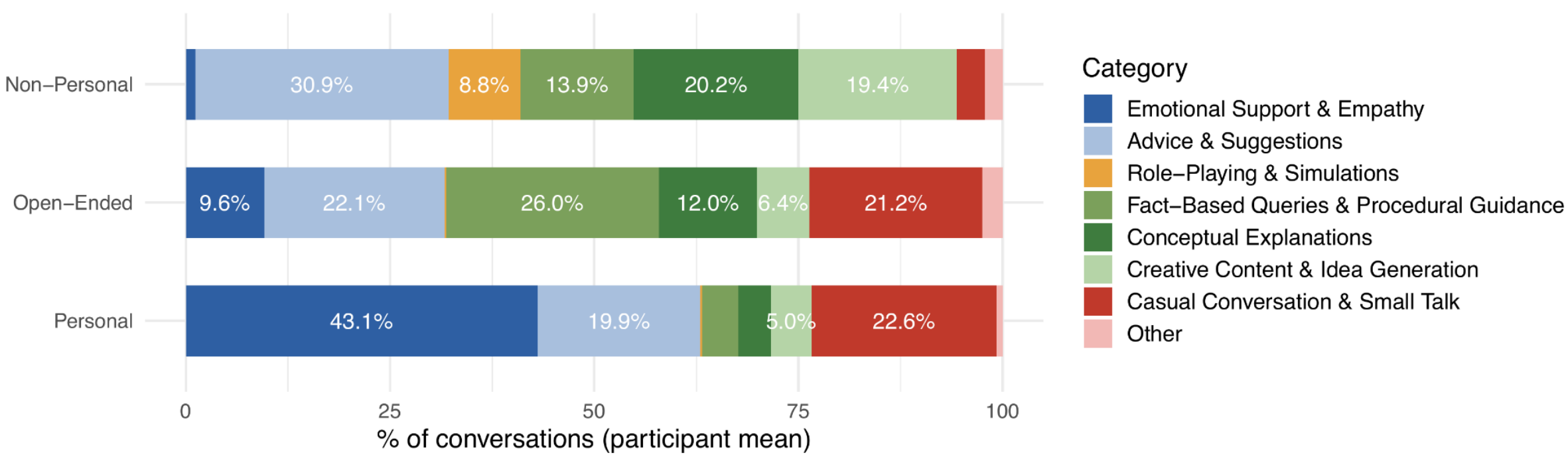

**Fig. 5 | Distribution of conversation topics by experimental condition.** Mean percentage of conversations in each topic category, averaged across participants within each condition. The Emotional Support & Empathy category accounted for 43.1% of conversations in the Personal condition, 9.6% in the Open-ended condition and a negligible share in the Non-personal condition.

**Did a month of daily AI use shift people's future preference toward AI for sharing personal matters?**

Across all participants, after interacting with AI for 28 days and regardless of whether they were assigned to have personal, non-personal or open-ended conversations, participants were asked again about their preferred partner for discussing personal issues. The proportion choosing AI as their preferred partner for personal issues increased from 25.4% to 31.9% ($t(980) = 3.88$; $P < 0.001$; $d = 0.12$). At the same time, the proportion choosing a human partner for the same scenario shifted in the opposite direction, declining from 82.1% to 76.5% ($t(980) = -3.67$; $P < 0.001$; $d = -0.12$).

Although we observed an overall change in people's preference, it was concentrated in the *Personal* and *Open-ended* conditions. Within-condition pre–post changes in AI preference for personal issues were +11.6% in *Personal* (31.6% → 43.2%), +9.5% in *Open-ended* (24.9% → 34.4%), and −1.2% in *Non-personal* (20.2% → 19.0%). The mirror pattern held for human preference: *Personal* (82.6% → 72.3%, −10.3%) and *Open-ended* (80.5% → 75.1%, −5.4%) showed meaningful declines, whereas *Non-personal* had virtually no change (83.1% → 81.6%, −1.5%). To test these differences statistically among the conditions, we fit a logistic regression predicting final AI or human preference from the *conversation topic* condition assignment, controlling for initial AI preference and *modality*. For AI preference, participants in both *Open-ended* (OR = 2.25, 95% CI, (1.55, 3.25), Tukey-adjusted P < 0.001) and *Personal* (OR = 3.03, 95% CI, (2.09, 4.38), Tukey-adjusted P < 0.001) conditions showed a significantly larger increase in their future preferences for AI compared to those in the *Non-personal* condition. For human preference, participants in the *Personal* condition showed a significantly larger decrease in preferring a human compared to those in the *Non-personal* condition (OR = 0.56, 95% CI, (0.38, 0.83), Tukey-adjusted P = 0.010); while those in the *Open-ended* condition showed a decrease in preferring a human but not significantly (Tukey-adjusted P = 0.15). See Fig. 6, Right.

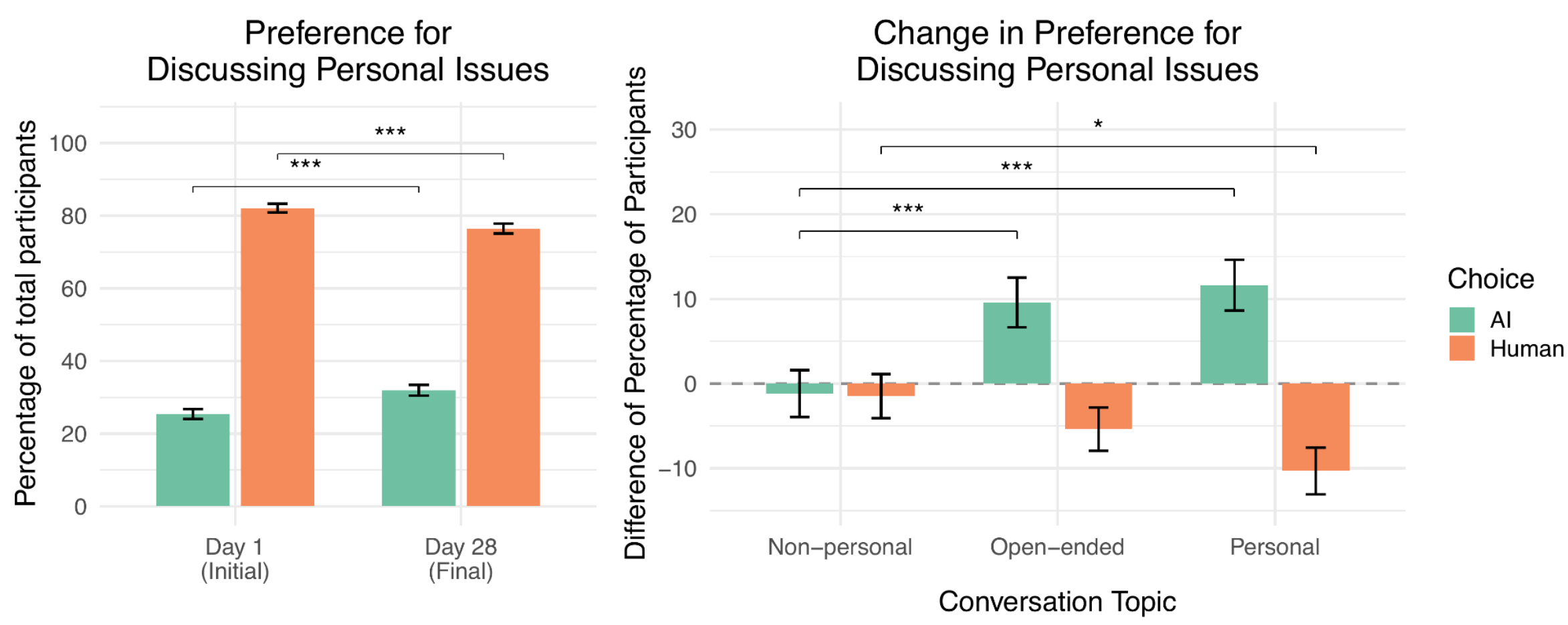

**Fig. 6 | Baseline and change in preferred partners for discussing personal issues.** (Left) Percentage of total participants choosing AI or a human as their preferred partner at the start and at the end of the study. (Right) Change in preference (Day 28 - Day 1, percentage points) for each conversation topic condition (Non-personal, Open-ended, Personal). *: P < 0.05, **: P < 0.01, ***: P < 0.001. P-values Tukey-adjusted for three pairwise comparisons. Error bars represent standard errors.

**What conversational behaviors predict future preference?**

To understand how specific conversational behaviors are associated with participants' future preferences for sharing personal issues with AI and humans, we analyzed the conversation corpus using a set of automated classifiers (i.e., LLM-as-a-judge[40]). The prompts of the classifiers can be found in Supplementary Information section 19, and the classifiers were validated by human annotators (see Supplementary Information section 19 for details).

We focused on 13 behaviors: seven behaviors capturing participants' affective language and self-disclosure (sharing problems, seeking support, expression of affection, attributing human qualities to the AI, and self-disclosure of personal information, thoughts, and feelings), and six behaviors capturing the AI's emotional engagement (personal questions, expression of affection, inquiry into personal information, empathetic responses, validation of feelings, and providing support resources). Each behavior was measured as the percentage of the total messages that contain the behavior for each participant, averaged across participants. Compared to the *Non-personal* and *Open-ended* conditions, the *Personal* condition showed a higher prevalence of emotionally engaging behaviors from both the AI and participants. See Supplementary Information section 16 for descriptive results.

We then examined whether these behaviors predicted participants' future preference for AI versus humans for discussing personal issues. For each behavior, we fit a logistic regression predicting post-study preference from the z-scored behavior rate, controlling for pre-study preference as well as conversation topic and modality, with Benjamini–Hochberg correction across the 13 tests per outcome (Fig. 7). Twelve of the 13 behaviors significantly predicted both outcomes — higher odds of choosing AI and lower odds of choosing a human. Among the AI's conversation behaviors (Fig. 7 top panel), validation of users' feelings (OR = 1.98, 95% CI, (1.59, 2.48), $P < 0.001$), expressing affection towards the user (OR = 1.80, 95% CI, (1.46, 2.21), P < 0.001) and expressing empathy (OR = 1.79, 95% CI, (1.40, 2.29), P < 0.001) predicted participants choosing AI (green lines), and the AI expressing empathy was also the strongest negative predictor of participants choosing a human (OR = 0.55, 95% CI, (0.42, 0.71), P < 0.001). Among the conversation behaviors of the participants (Fig. 7 bottom panel), sharing problems (OR = 1.91, 95% CI, (1.55, 2.34), P < 0.001), seeking support (OR = 1.85, 95% CI, (1.49, 2.30), P < 0.001), and self-disclosure of feelings (OR = 1.80, 95% CI, (1.48, 2.19), P < 0.001) strongly predicted future preference for AI while negatively predicting choosing a human (all OR < 1, P < 0.01). Only the AI's inquiry into personal information predicted neither outcome (both P > 0.13, corrected). These results showed that emotionally engaging behaviors in the conversations with AI matter for shifting people's future preferences on who to talk about personal issues with.

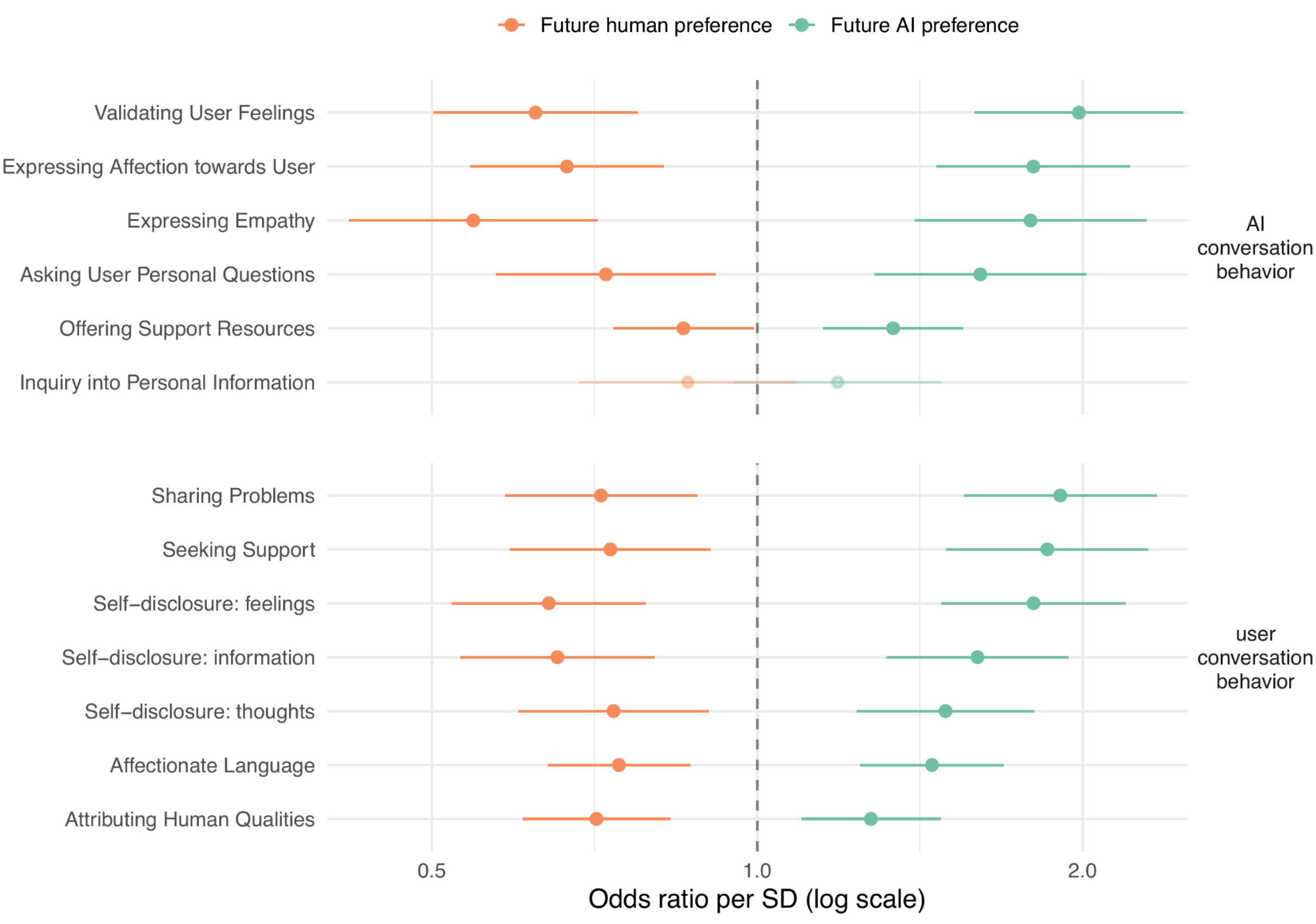

**Fig. 7 | Conversation behaviors predict future preference for discussing personal issues.** Odds ratios per SD of each behavior's rate from logistic regressions predicting post-study

preference — choosing AI (green) and choosing a human (orange) — controlling for pre-study preference, conversation topic, and modality. Points right of the dashed line (OR = 1) indicate higher odds of that preference per 1 SD increase in the behavior; horizontal lines are 95% CIs (log scale). Filled points are significant after Benjamini–Hochberg correction across the 13 tests within each outcome.

**Discussion**

The current research examined the psychological drivers of choosing AI for emotional support, how choice shapes people's experience of the interaction, and how exposure to AI emotional support shapes subsequent choices. Across three lab studies and a month-long field study conducted in collaboration with OpenAI, three main findings emerged. First, in exploring drivers of choice, we found large differences in beliefs between participants who chose humans and those who chose AI. Participants who preferred to share their emotions with humans believed humans were better at providing emotional support than AI and that humans and AI were similar in how nonjudgmental they were perceived to be. In contrast, participants who preferred to share their emotions with AI believed that humans and AI were equally capable of providing emotional support, but that AI was much more nonjudgmental than humans. These results suggest that people hold very different beliefs about the nature of AI emotional support, and that these beliefs shape their choice of emotional support partner.

In exploring the consequences of emotional support choice on participants' experiences, we found that AI was rated higher than humans only when people had initially preferred and had been assigned to AI. This finding suggests that the AI superiority in emotional support documented in the literature is not a general effect, but is concentrated among participants who initially preferred AI. As for the consequences of interacting with AI on future preferences, we found that, regardless of people's initial choice, interacting with AI about emotional or personal experiences updated people's beliefs about AI's capacity for emotional support and increased their likelihood of choosing AI over a human for future sharing. Our longitudinal study extended this finding to a naturalistic setting. Participants were asked to interact with ChatGPT for at least five minutes a day, talking about either non-personal issues, personal issues, or open-ended topics. We found that both the open-ended and personal-issue conversations led to an increase in participants' willingness to use AI to discuss personal issues and a decrease in their likelihood of choosing a human. Overall, these results reveal the importance of choice in emotional support interactions.

These results have important implications for our understanding of emotional support choice. A growing body of research suggests that AI is superior to humans in providing short messages of emotional support[19]. Our studies isolate choice as an important moderator of AI superiority in short empathic messages as AI superiority is seen only when participants actually chose to interact with AI. Furthermore, regardless of whether participants chose a human or an AI partner, our results point to an important consequence of choice: it is highly path-dependent, such that previous engagement substantially influences future choice. Participants were much more likely to choose whichever partner they were initially assigned to in subsequent choices. In our lab studies, 70.1% of participants assigned to chat with an AI later chose AI again, compared to 45.5% among those assigned to a human. This is especially consequential for AI emotional support, which is readily available at any time[1,22]. Our longitudinal study supported this finding: 5-minute interactions with AI over 28 days led to an increase in AI preference for personal conversations from 25.4% to 31.9% and a decrease in human preference from 82.1% to 76.5%. As users "stumble into" AI emotional support in everyday life, their preferences for a future partner shift, and the probability of choosing AI in future situations increases.

As sharing emotional and personal experiences is central to building and maintaining relationships, the path dependence in choice captured by our studies could shift humans toward seeking AI emotional support as a result of AI availability. It is hard to tell the limits of the path dependence in AI emotional support. It is possible that AI will shift preferences to a certain extent, but people will largely still prefer to interact with other humans. It is also possible that AI emotional support will shift people's choices, until some people never prefer human connection. Either way, our results point to important design opportunities. When an AI system detects that a user is engaging in emotional interactions, it could remind and encourage that user to reach out to humans as well. Encouraging contact with supportive humans in people's lives may mitigate the path-dependent shift toward AI.

**Limitations and future directions**

There are several limitations of this research. First, we compared AI only with a human stranger in our lab studies, but in real life the alternative to AI emotional support is often a friend, romantic partner, family member, or professional therapist. We acknowledge that the results may differ when participants are asked to choose between AI and a friend, as rich lived experience matters for both perceived and experienced emotional support to be "thick" and meaningful[41]. Study 4 was designed to partially address this limitation. Our preference measure did not specifically refer to a human stranger, yet we did observe that participants were more likely to choose to share personal experiences with an AI after the 28 days. Nevertheless, future research should incorporate a broader range of human options to examine how people choose when close others or professionals are available and how people experience these interactions compared to AI.

Second, in everyday life, people may encounter AI emotional support through many pathways other than deliberately seeking support from AI. Recent conceptual and empirical work suggests that emotional support often emerges incidentally during task-oriented use of general-purpose AI platforms[37,38]. It is therefore unclear whether the effects we observed in more structured settings generalize across the diverse real-world scenarios in which people receive AI emotional support. Study 4 provides some initial evidence on these broader pathways by asking participants to interact with AI across a wider set of topics (i.e., personal, non-personal, and open-ended) in daily life for at least five minutes every day. The open-ended conversation condition, which was the closest to typical everyday use, showed that daily AI use over one month increased preference for talking about personal matters with AI and decreased preference for human partners. Future research should further test these results in naturalistic contexts to understand how people experience emotional support, how choice may influence it, and how their choices about whom to approach for support may evolve over time.

Third, our measures of preference change relied on self-reports rather than actual behaviors on whom participants turned to for support in their daily lives. It remains an open question whether these reported preferences translate into changes in real support seeking behavior. Future work should track actual behaviors of support partners in real-world settings and, crucially, examine the downstream consequences for how people navigate social relationships and how these shifts affect well-being over time.

## Methods

### Ethical statement

Studies 1-3 were approved by the Institutional Review Board at Harvard University (IRB24-1718 and IRB25-0035). OpenAI and MIT jointly obtained Institutional Review Board (IRB) approval for Study 4 through Western Clinical Group (WCG) IRB (#20243987). Informed consent was obtained from participants in all studies.

**Preregistrations**

The studies were all preregistered (Study 1: https://osf.io/azjux/files/t9rk2?view_only=3f026921a99e4714847a8a290aa36c1f; Study 2: https://osf.io/azjux/files/x5ecg?view_only=3f026921a99e4714847a8a290aa36c1f; Study 3: https://osf.io/9hsyc/overview?view_only=cca34e8db5da4db9b7f8092c9e2a6cf3; Study 4: aspredicted.org/7xhy-ds3c.pdf).

**Participants**

**Study 1.** As preregistered, we recruited 250 participants (Mean age = 40.1, SD = 15.0; 63.1% female; 77.9% White) from Prolific in March 2025, each of whom was compensated $1.80. After excluding 20 participants who failed the attention check question and 1 participant who did not consent to participate in the study, the final sample included 229 participants. We initially preregistered this sample size to test the effect of emotional valence on choice and estimated the sample based on this question. We assumed a medium effect size (odds ratio of the logistic regression = 1.5) and base probability = 0.3 (probability of sharing with AI when emotion valence is at the mean) with 80% power and 0.05 significance-level for the sample size calculation.

**Study 2 and Study 3.** Based on Study 1 results, we expected ~75% of participants would choose to interact with an AI in this design, and therefore that the two conditions involving participants who chose a human partner would yield the smallest samples. Our total sample size was therefore determined by these conditions. A power analysis indicated that detecting a medium effect (Cohen's $d = 0.5$) with 80% power required 64 participants per condition; therefore, we needed a total of 512 participants. In the second part of the survey, half of the participants in the human condition would be assigned to a "listener" role and not report their own emotional experience, which requires an additional 256 participants. To accommodate potential matching failures and other preregistered exclusions, we targeted and preregistered a final recruitment of 850 participants.

All participants of Studies 2 and 3 were recruited through the CloudResearch Connect platform using prescreening criteria requiring U.S. residency, fluent English language status, and an approval rate of at least 90%. Because the study required real-time dyadic interaction, we needed to ensure a high and sustained influx of participants during recruitment so that participants in the human condition could be paired promptly. Therefore, we recruited participants in batches (Study 2: from September 23 to October 3, 2025; Study 3: from March 18 to April 2, 2026). To increase the completion rate, we designed the survey so that participants assigned to the human condition were automatically switched to the AI condition if they were not matched within 3 minutes. This switch was designed to happen before they were informed of their actual partner. After each batch of data collection, we monitored the distribution of participants across conditions. During Study 2 recruitment, there was a technical error. When the study reached N = 600, the counts (after accounting for participants who switched from human to AI due to matching delays) were 269 participants in the AI condition and 215 participants in the sharer role of the human condition. To reduce this imbalance, the following batch of 100 participants (50 of which were sharers) was assigned only to the human condition, which brought the totals to 269 in AI and 265 in human. Therefore, for the final batch of 50 participants, we reinstated random assignment. However, when adding back the AI condition in the survey flow, we inadvertently left the "evenly present elements" option enabled and did not reset the condition counters. Because the AI condition had a count of zero at that moment, the survey randomizer assigned all 50 participants in that batch to the AI condition. Given our preregistered sample size of 850, we had to correct the imbalance by recruiting an additional 34 participants into the human condition and randomly discarding 34 observations from that misassigned batch so that the final dataset reflected the intended allocation.

Study 2 sample consisted of 851 participants (Mean age = 41.0, SD = 11.7; 53.2% female; 69.7% White), each of whom was compensated $3.40. As preregistered, we excluded 2 participants who failed the attention check and 7 participants who failed the CAPTCHA check (threshold < 0.5), resulting in a sample of 842 participants. Of these, 287 were assigned to the AI partner condition, and 555 were assigned to the human partner condition (including both sharer and listener roles). For the post-conversation analyses, as preregistered, we applied the following exclusion criteria to ensure the interaction quality. First, we excluded 27 participants who failed to successfully interact with a human partner due to partner dropout after being matched. We then excluded 40 participants who explicitly reported experiencing technical difficulties during the conversation (i.e., 35 participants in the human condition and 5 participants in the AI condition). We also excluded 69 participants for whom either they or their partner sent fewer than three messages during the conversation (53 participants in the human condition and 16 participants in the AI condition). Because our primary analysis focused on participants in the sharer role, we additionally excluded all listeners from the human-partner condition. After applying all preregistered exclusions, the final sample consisted of 487 participants (Mean age = 40.37, SD = 11.2; 54.6% female; 67.8% White): 221 in the human condition and 266 in the AI condition. 183 participants chose an AI partner and chatted with an AI; 149 chose an AI partner but chatted with a human; 83 chose a human partner but chatted with an AI; and 72 chose and chatted with a human.

Study 3 recruited 850 participants (Mean age = 41.2, SD = 13.3; 59.5% female; 72.0% White), each of whom was compensated $3.40. As preregistered, we excluded 7 participants who failed the attention check and 20 participants who failed the CAPTCHA

check (threshold < 0.5), resulting in a sample of 823 participants. Of these, 309 were assigned to the AI partner condition, and 514 were assigned to the human partner condition (including both sharer and listener roles). For the post-conversation analyses, we applied the same preregistered exclusion criteria as in Study 2. We excluded 24 participants who failed to successfully interact with a human partner due to partner dropout after being matched, 51 participants who reported technical difficulties during the conversation (34 in the human condition and 17 in the AI condition), and 100 participants for whom either they or their partner sent fewer than three messages during the conversation (71 in the human condition and 29 in the AI condition). Because our primary analysis focused on participants in the sharer role, we additionally excluded all listeners from the human-partner condition. After applying all preregistered exclusions, the final sample consisted of 457 participants: 194 in the human condition and 263 in the AI condition. Among these participants, 175 chose an AI partner and chatted with an AI; 109 chose an AI partner but chatted with a human; 88 chose a human partner but chatted with an AI; and 85 chose and chatted with a human. For the pre-conversation analysis, including all participants did not substantially change the results, and for the post-conversation analysis, including all participants who had the conversation with their assigned partner did not substantially change the results.

**Study 4.** A total of 2,539 participants were enrolled in the study through CloudResearch. 1,911 completed the pre-study survey and 1,394 completed the four-week protocol. As preregistered, we excluded 644 for missing daily tasks (e.g., <3 daily tasks / week); 101 for low chatbot usage (<10 messages/session or <12 sessions total); 12 for no recorded chatbot usage; 535 for missing pre- or post-survey; 102 for missed weekly surveys; 29 for being assigned voice-conditions but interacted with the chatbot via text; 13 for incomplete survey records or a self-reported gender outside the two categories (male and female) the analysis models accommodated. 122 participants either withdrew or were removed for duplicated entries.

The final set of participants consisted of 981 people with a mean age of 39.9 (SD = 11.6) and an almost equal split of male and female (Female: 510, 52.0%, Male: 471, 48.0%). Most participants identified as White (734, 74.8%), and 126 (12.8%) as Black or African American, 65 (6.6%) as Asian. About half (463, 47.2%) had used the text modality of ChatGPT at least a few times a week. Most (683, 69.6%) had never used the voice modality of ChatGPT, likely because voice is a new feature. About a third (344, 35.1%) had used other assistant-type chatbots more than a few times a week (e.g., Google's Gemini, Anthropic's Claude), and most (696, 70.9%) had never used companion chatbots (e.g., Replika, Character.ai). The full demographic breakdown is in Supplementary Information section 21.

**Procedure**

**Studies 1-3.** In Studies 1-3, participants first completed an attention check question designed to ensure they had read the instructions carefully (in Studies 2 and 3, participants completed a CAPTCHA question before the attention check question), and then they were instructed to think about a significant emotional experience they had within the past month, which could involve situations in either their personal or professional lives and remained vivid and easy to recall. They then described this experience in a free-response text box, writing 4–5 sentences about what happened, where they were, who else was involved, what emotions they felt, and how the experience unfolded. Following the description, participants rated the emotional valence of the experience and indicated the intensity of eight discrete emotions (i.e., happiness, sadness, fear, anger, excitement, pride, anxiety, and enthusiasm). Participants were then informed that they would later engage in a brief text-based conversation about the emotional experience they had described, with either another participant or an AI (a GPT-based or Claude-based chatbot). In Study 1, participants first chose who they would like to share their emotional experience with (using both a binary choice question and continuous rating scales for each option) and were asked to provide explanations for both why they selected their chosen option and why they did not select the alternative in open-ended text boxes. They then rated their perceptions of sharing their emotional experience with a human and an AI using 14 items (See Main Dependent Variables section for specific items). In Studies 2 and 3, we reversed the order by first asking participants to rate their perceptions of sharing their emotional experience with a human and an AI on the same 14 items (measuring four perception dimensions: emotional support, judgment concerns, confidentiality, and advice quality). This was to rule out the possibility that the ratings reflected motivated perceptions after the choice. They then rated the importance of these four factors and indicated which partner they would prefer to talk to about their emotional experience. Participants in Study 1 then reported their general attitudes toward generative AI, how often they use generative AI, their familiarity with generative AI, and rated their frequencies of sharing emotions with AI and humans. Finally, they provided demographic information. At the end of Study 1, participants were debriefed that they would not have a direct conversation with a partner in this study, but would receive a response from their selected partner in a few days (the human responses were written by the first author and sent to the participants who chose human, and the AI responses were generated by GPT-4o and sent to the participants who chose AI).

In Studies 2 and 3, after completing the above measures about expectations and preferences, participants entered a pairing phase supported by SMARTRIQS[42]. They were randomly assigned to one of two conditions: (1) an AI condition, in which they were directed to chat with a chatbot (powered by GPT-4o in Study 2 and powered by Claude Opus 4.6 in Study 3) prompted to act as an empathetic listener, or (2) a human partner condition, in which they were matched with another participant completing the study simultaneously but who was assigned to an empathetic listener role. The prompt for the AI chatbot was matched in content to the instructions given to the listener in the human condition. The prompt reads: "You will have a friendly 5-minute conversation with a human partner about their emotional experience. In the conversation, you will act as an empathetic listener. During the conversation, your partner (i.e., a human) will share their emotional experience with you. Your role is to read their experience and offer supportive, empathetic responses throughout the conversation. Do not include any Emoji in your message. Each of your messages should not be very long, keep it no more than three or four sentences. Each message should be in one paragraph. You can acknowledge and validate their feelings using warm and understanding language, and ask open-ended questions to help them reflect and explore if appropriate. Please present yourself as a neutral and supportive conversation partner, and try to have a thoughtful and meaningful chat with your partner." Participants were explicitly informed about whom they were assigned to and that no deception was used. For participants assigned to the AI condition, they were instructed to chat with the AI chatbot through the LEXI platform[43] embedded in the survey. For participants assigned to the human condition, they first entered a real-time matching queue. Once matched, pairs were randomly assigned to one of two conversational roles: Sharer (who would describe the emotional experience they recalled) or Listener (who would provide supportive, empathetic responses). Participants received role-specific instructions and then engaged in a 5-minute text-based conversation. The interface of the chat window was similar for both conditions. The structure of the conversation was also the same: participants began by sharing the emotional experience they recalled, followed by exchanges of text messages with their assigned partner until the chat reached 5 minutes.

Across both studies, conversations in the AI condition involved an average of 12.2 exchanges of messages (SD = 4.69), while those in the human condition involved 11.4 exchanges (SD = 4.02; Study 2: $M_{AI}$ = 13.18, SD = 4.85, $M_{Human}$ = 11.64, SD = 3.99; Study 3: $M_{AI}$ = 11.20, SD = 4.31, $M_{Human}$ = 11.07, SD = 3.98). In terms of word count per turn, participants who shared with a human listener produced an average of 20.96 words per turn (SD = 10.34; Study 2: M = 20.53, SD = 10.75; Study 3: M = 21.45, SD = 10.01), whereas those who shared with the AI produced 28.66 words per turn (SD = 16.75; Study 2: M = 26.86, SD = 16.53; Study 3: M = 30.48, SD = 16.80). On the listener side, human listeners produced an average of 14.67 words per turn (SD = 5.78; Study 2: M = 14.35, SD = 6.04; Study 3: M = 15.03, SD = 5.98), while the AI chatbot produced substantially more, averaging 56.63 words per turn (SD = 14.85; Study 2: M = 47.28, SD = 9.74; Study 3: M = 66.08, SD = 13.05).

Immediately after the conversation, participants completed post-conversation measures, reporting their perceptions of their assigned partner on the four dimensions with the same items as the pre-conversation measures: emotional support, nonjudgment, confidentiality, and advice quality. Participants also rated the outcomes of the conversation, their enjoyment and satisfaction with the conversation, their desire to continue the interaction, their current emotional state (i.e., positivity and negativity), a binary choice question about whom they would prefer to talk to about a significant emotional experience in the future (a human stranger or an AI companion), and two continuous items assessing the extent to which they would want to share emotions with each type of partner. Finally, as in Study 1, participants completed additional measures assessing their general attitudes toward generative AI, how often they use generative AI, and their familiarity with generative AI. They also rated their frequencies of both sharing emotions with AI and humans. They then provided demographic information and reported any technical difficulties they experienced during the study.

**Study 4.** We conducted a one-month randomized controlled trial (n = 981, 28 days, > 300,000 messages) that crossed three interaction modes ("Modality": text only, a neutral and professional voice, or an engaging and expressive voice) with three conversation types ("Conversation topic": open-ended, non-personal or personal conversation prompts) in a 3 × 3 factorial design. Participants were asked to use OpenAI's ChatGPT (GPT-4o) for at least five minutes daily and were randomly assigned to one of nine conditions. At the start of the study, each participant completed a pre-study survey that established baseline measures for their initial preference for their partner (AI vs human vs no preference) for different scenarios as well as the participants' prior characteristics. Throughout the study, participants received daily emails with a daily survey containing specific prompts they were to discuss with the AI model. These prompts were aligned with their assigned conversation topic category (open-ended, non-personal, or personal conversation). Participants were asked to interact with the chatbot for minimally five minutes, with no limits beyond the required usage duration. Participants completed three weekly surveys and, at the fourth week, a post-study survey and followed an off-boarding protocol. The weekly and post-study surveys captured changes in the dependent variables relative to baseline measures.

**Main Dependent Variables.**

**Beliefs of Emotional Sharing with Human and AI.** Participants completed a 14-item measure assessing perceptions of sharing emotional experiences with a partner. After the interaction, they rated their assigned partner again on these items. This design helps us to directly compare their pre-conversation beliefs with their actual experience with the partner. The items are grouped into four perception dimensions. The specific items are: 1) Perceived judgment (AI $\alpha$ = 0.75; Human $\alpha$ = 0.80; Post-conversation $\alpha$ = 0.68): I can share the emotional experience in a non-judgmental setting; I will be concerned about my impression (reverse-coded); I can be true to myself; I can be open and honest about my emotions; 2) Perceived emotional support (AI $\alpha$ = 0.93; Human $\alpha$ = 0.92; Post-conversation $\alpha$ = 0.93): I will feel heard and validated; I will feel understood; I will feel emotionally supported; I will receive empathetic responses; I will feel socially connected; I will feel close to the partner; 3) Perceived advice (AI $\alpha$ = 0.86; Human $\alpha$ = 0.83; Post-conversation $\alpha$ = 0.83): I will receive useful advice; I will gain new insights about my emotions; 4) Perceived confidentiality (AI $\alpha$ = 0.75; Human $\alpha$ = 0.64; Post-conversation $\alpha$ = 0.63): I can trust that my experience will be kept confidential; I will worry about my privacy (reverse-coded). All items were answered on a 7-point scale (1 = strongly disagree, 7 = strongly agree). In Studies 2 and 3, pre-interaction measurement items referred to both an AI partner and a human stranger, and post-interaction items referred to the assigned conversational partner using past-tense wording.

**Partner Choice.** In Studies 1-3, participants indicated which partner they would prefer to talk to about their emotional experience. Participants were informed that they would later speak with either another human participant or an AI companion (an AI-powered chatbot). They were then asked: "If you could choose, who would you prefer to talk to about your emotional experience?" and selected either an AI companion or another participant. The conversation was described as anonymous and confidential to elicit participants' true preferences. Participants also provided continuous ratings of how much they would like to talk to the types of partners. The wording of these questions differed slightly across studies. In Study 1, participants rated how comfortable they felt to have a conversation about the emotional experience they recalled with an AI companion and another participant from 0 (Not comfortable at all) to 10 (Very comfortable). In Studies 2 and 3, participants rated how much they would like to talk about the emotional experience they recalled with an AI companion and another participant from 1 (Not at all) to 7 (Very much).

**Experienced Interaction Quality.** In Studies 2 and 3, following the interaction, participants reported their experienced quality of the interaction, including how satisfied they were with the conversation, how much they enjoyed it, and how much they would like to continue the conversation with their partner. All three items were rated from 1 (Not at all) to 7 (Very much).

**Future Choice.** In Studies 2 and 3, participants were asked: “Next time you have a significant emotional experience, between the below two options, who would you like to talk about the emotional experience with?”, choosing between a human stranger and an AI companion. Participants also provided continuous ratings of how much they would like to talk about a future emotional experience with a human stranger and with an AI companion, each on the same 1–7 scale. In Study 4, participants were asked about their choice for interacting with a human or a chatbot across seven different scenarios: “casual conversation”, “asking for information”, “asking for help with a task”, “venting/talking about personal issues”, “talking about past events / reviewing events in a day”, “talking about plans for the future”, and “entertainment and play”. We focus on “personal issues” as it is most relevant to our investigation. Participants could choose from “chatbot”, “human, online”, “human, in-person”, and “no preference”, and they could choose multiple options for each situation. These questions were asked both before (at day = 1) and after the study (at day = 28). We later joined the responses that included either of the two “human” options (i.e., the human option is counted as chosen if the “human, online” or “human, in-person” was selected and only counted once if both were selected). We then binarized the choices: choosing AI vs not choosing AI, and choosing human vs not choosing human.

**Other Measures.** In all Studies 1-3, we measured participants' typical emotional-sharing behavior with AI and with humans. Participants indicated how often they usually share their emotional experiences with an AI companion (*"How often do you usually share your emotional experiences with AI companions (i.e., an AI-powered chatbot)?"*) and with other people (*"How often do you usually share your emotional experiences with other people (e.g., your friends, family members, strangers)?"*), each using the same 0–10 scale (0 = Not at all, 10 = All the time).

## Data availability

All preprocessed data, excluding participants' emotional experiences and chat messages, are available via OSF at https://osf.io/q7r2w/overview?view_only=bed0c4380c9a4bfdb5c5bd3642cdacda.

## Code availability

All analysis files are available via OSF at https://osf.io/q7r2w/overview?view_only=bed0c4380c9a4bfdb5c5bd3642cdacda.

## Author Contributions

Y.S., C.M.F., and A.G. designed research, conducted experiments, analyzed data, and wrote the manuscript. Y.S., C.M.F., G.L., P.M., and A.G. all contributed to the data collection, reviewed and edited the manuscript.

## Competing Interests

The authors declare no competing interests.